 \documentclass[pmlr,twocolumn,10pt]{jmlr} % W&CP article

\usepackage{booktabs}
\usepackage{siunitx}

\theorembodyfont{\upshape}
\theoremheaderfont{\scshape}
\theorempostheader{:}
\theoremsep{\newline}

\title[Augmentations for Health Acoustic Signals]{Optimizing Audio Augmentations for Contrastive Learning of Health-Related Acoustic Signals}

\author{%
\Name{Louis Blankemeier} \Email{blankemeier@google.com}\\
\Name{Sebastien Baur} \Email{sebastienbaur@google.com}\\
\Name{Wei-Hung Weng} \Email{ckbjimmy@google.com}\\
\Name{Jake Garrison} \Email{jakegarrison@google.com}\\
\Name{Yossi Matias} \Email{yossi@google.com}\\
\Name{Shruthi Prabhakara} \Email{shruthip@google.com}\\
\Name{Diego Ardila} \Email{ardila@google.com}\\
\Name{Zaid Nabulsi} \Email{znabulsi@google.com}\\
\addr Google Research, USA
}
\begin{document}

\maketitle

\begin{abstract}
Health-related acoustic signals, such as cough and breathing sounds, are relevant for medical diagnosis and continuous health monitoring. Most existing machine learning approaches for health acoustics are trained and evaluated on specific tasks, limiting their generalizability across various healthcare applications. In this paper, we leverage a self-supervised learning framework, SimCLR with a Slowfast NFNet backbone, for contrastive learning of health acoustics. A crucial aspect of optimizing Slowfast NFNet for this application lies in identifying effective audio augmentations. We conduct an in-depth analysis of various audio augmentation strategies and demonstrate that an appropriate augmentation strategy enhances the performance of the Slowfast NFNet audio encoder across a diverse set of health acoustic tasks. Our findings reveal that when augmentations are combined, they can produce synergistic effects that exceed the benefits seen when each is applied individually.
\end{abstract}

\section{Introduction}
\label{sec:intro}
Non-speech, non-semantic sounds, like coughing and breathing, can provide information for doctors to detect various respiratory diseases, cardiovascular diseases and neurological diseases~\citep{boschi2017connected,zimmer2022making}. Advances in deep learning-based machine learning (ML) allow us to develop medical assistants and continuous health monitoring applications by learning effective acoustic data representations~\citep{alqudaihi2021cough}.

Current approaches for learning health acoustic representations are mostly trained and evaluated on specific tasks. For example,~\cite{botha2018detection,larson2012validation,tracey2011cough,pahar2021automatic} trained models to detect tuberculosis using cough sounds via supervised learning. However, it can be challenging to adopt these models directly for other health acoustic tasks. Retraining task specific health acoustic models requires manual data collection and labeling by clinical experts, which can be time consuming and costly. 

Researchers within the ML community have explored various self-supervised strategies to learn general purpose data representations that overcome the limitations of domain-specific representations~\citep{balestriero2023cookbook}. Among these approaches, contrastive learning has proven effective for generating robust representations across multiple data modalities, including images, videos, speech, audio, and periodic data~\citep{chen2020simple,jiang2020speech,qian2021spatiotemporal,oord2018representation,yang2022simper}. Selecting appropriate data augmentations is crucial for performant contrasting learning algorithms~\citep{chen2020simple} (see Related Works for details). Consequently, significant research has been conducted on the utility of various augmentations for images~\citep{chen2020simple}, videos~\citep{qian2021spatiotemporal}, and speech/audio~\citep{al2021clar,jiang2020speech}. However, the unique characteristics of health-related acoustic signals, such as coughs and breathing sounds, which differ in pitch and tone from speech and music, raise questions about the applicability of existing contrastive learning and augmentation strategies in this specialized domain.

To address this research gap, our study systematically explores eight distinct audio augmentation techniques and their combinations in the context of health acoustic representation learning. We employ the self-supervised contrastive learning framework, SimCLR~\citep{chen2020simple}, with a Slowfast NFNet backbone~\citep{wang2022towards}. After identifying the best combination of augmentations, we compare the performance of the resulting Slowfast NFNet against other state-of-the-art off-the-shelf audio encoders on 21 unique binary classification tasks across five datasets. This work offers two major contributions: (1) we identify augmentation parameters that work best when applied to health acoustics, and (2) we investigate the synergistic effects of combining audio augmentations for enhancing health acoustic representations using SimCLR.

\section{Related Works}
\label{sec:related}
In ML, data augmentation serves as a regularization technique to mitigate the risk of model overfitting~\citep{zhang2021understanding}. Within the framework of contrastive learning, the objective is to learn data representations that minimize the distance between representations of semantically similar inputs and maximize the distance between representations of semantically dissimilar inputs. Data augmentations are critical for contrastive learning-based self-supervised learning (SSL), and eliminates the need for labeled data for representation learning. By applying a variety of augmentations to a single input, semantically consistent but distinct variations, commonly referred to as views, are generated~\citep{von2021self}. The task then becomes pulling these related views closer together in the representational space, while concurrently pushing views derived from different, unrelated inputs farther apart, via a contrastive loss, such as InfoNCE in SimCLR~\citep{chen2020simple}. This approach establishes a form of invariance in the model, rendering it robust to the augmentations applied during the training process.
Augmentations have been widely explored as part of contrastive learning-based SSL methods such as SimCLR, BYOL~\citep{grill2020bootstrap}, MoCo~\citep{chen2020improved}, and SwAV~\citep{caron2020unsupervised}. Data augmentations also enhance the performance of SSL methods broadly across different data modalities, including images~\citep{chen2020simple}, videos~\citep{qian2021spatiotemporal}, audio~\citep{al2021clar,niizumi2021byol}, speech~\citep{jiang2020speech}, and 1-dimensional signals (e.g., human physiological signals)~\citep{yang2022simper}. In this study, we turn our attention toward a relatively underexplored domain: the application of data augmentations strategies for contrastive learning of health acoustic signals.

The most closely related area of research to our focus on health acoustics is the research investigating augmentation strategies for speech and audio data. Early research by~\cite{ko2015audio} explored creating two augmented speech signals with speeds relative to the original of 0.9 and 1.1. This yielded performance improvements across four speech recognition tasks.~\cite{jansen2018unsupervised} expanded upon this by introducing a triplet loss for audio representation learning, incorporating random noise, time/frequency translation, example mixing, and temporal proximity augmentations.~\cite{jiang2020speech} employed an adaptation of SimCLR for speech data, termed Speech SimCLR, where they applied a diverse set of augmentations: random pitch shift, speed perturbation, room reverberation and additive noise to the original waveform, as well as time and frequency masking to the spectrogram.~\cite{niizumi2021byol} developed a comprehensive audio augmentation module including pre-normalization, foreground acoustic event mixup, random resize cropping and post-normalization.~\cite{fonseca2021unsupervised} investigated a multi-modal approach by adopting augmentations from both vision and audio domains, including random resized cropping, random time/frequency shifts, compression, SpecAugment~\citep{park2019specaugment}, Gaussian noise addition, and Gaussian blurring. They also used sound separation techniques for sound event detection to enable targeted data augmentations~\citep{fonseca2021self}.~\cite{shi2022robust} explored the impact of noise injection as an augmentation strategy to bolster the robustness of speech models. CLAR identified six augmentation operations: pitch shift, noise injection in frequency domain, and fade in/out, time masking, time shift, time stretching in the temporal domain, and explored their utility for audio contrastive learning~\citep{al2021clar}. In this study, we build upon these ideas to systematically investigate the optimal combination and sequence of augmentation strategies, with a specific focus on developing robust representations for health acoustics.

\section{Methods}
\label{sec:methods}
The study is structured into three phases. The first phase consists of finding the best parameters for each augmentation that we consider for use with SimCLR. In the second phase, we investigate various combinations of augmentations, where we apply one or two successive augmentations to create each view of the input. Here, we use the augmentation parameters that we select in the first phase. In the third phase, we compare the results of our best performing model to other state-of-the-art audio encoder models on the validation set used for comparing augmentations. We choose to hold out the test sets due to ongoing model development and these results may thus be optimistic. This evaluation involves 21 unique downstream tasks across five datasets and we investigate the quality of embeddings generated from each audio encoder using linear probing~\citep{kohn2015s}. Our study employs SimCLR with a 63 million parameter SlowFast NFNet-F0 as the neural network backbone~\citep{chen2020simple,wang2022towards}.

\begin{figure}[h]
\centering
\includegraphics[width=1.0\linewidth]{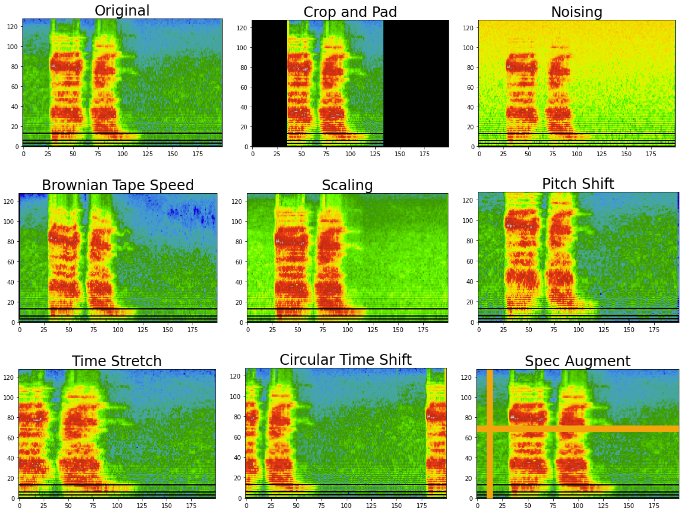} 
\caption{Mel spectrograms generated from various augmentations applied to the same health acoustic sample. One two-second example from the CoughVID dataset~\citep{orlandic2021coughvid} is acquired and modified by each augmentation method.}
\label{fig1}
\end{figure}

\paragraph{Audio Augmentations}
We investigate eight augmentations (Figure~\ref{fig1}). These include the following time-domain augmentations: crop and pad, noising, Brownian tape speed~\citep{weng2023predicting}, scaling, pitch shift, time stretch, and circular time shift. Additionally, we experiment with SpecAugment which is applied after the transformation of audio inputs into spectrograms~\citep{park2019specaugment}. A description of each augmentation strategy is provided in Appendix Table A1.

Each of the augmentations offers a tunable parameter space to allow for varying degrees of transformational intensity. To identify the optimal hyperparameters for each specific augmentation, we first conduct an exhaustive grid search. After we determine the best augmentation parameters, we explore the potential synergistic effects from the sequential application of either one or two successive augmentations. Since we include 8 augmentations, experimenting with every permutation of one or two augmentations would result in 64 experiments. However, in this work, SpecAugment was only applied after the time domain augmentations which reduced the number of 2-step augmentations to 57. 

\paragraph{Datasets}
For this study, we curate a training dataset, YT-NS (YouTube Non-Semantic), consisting of two-second long audio clips extracted from one billion non-copyrighted YouTube videos, totalling about 255 million 2s clips or 142k hours. We apply a convolutional neural network-based health acoustic detector model, trained on two public health acoustic AudioSet derivatives, FSD50K and Flusense, as well as another health acoustic dataset. We use this model to filter two-second audio clips from these one billion videos for the following health acoustic signals: coughs, speech, laughing, throat-clearing, baby coughs, and breathing. Estimated numbers of each of these clips is provided in Appendix Table A2. The Slowfast NFNet encoder is trained solely using this dataset.

For evaluation, we use five publicly available datasets, FSD50K~\citep{fonseca2021fsd50k}, Flusense~\citep{al2020flusense},~\href{https://www.scidb.cn/en/detail?dataSetId=778740145531650048&dataSetType=personal}{PSG}~\citep{korompili2021psg},~\href{https://zenodo.org/record/4498364#.Yf21afXMJqv}{CoughVID}~\citep{orlandic2021coughvid}, and~\href{https://github.com/iiscleap/Coswara-Data}{Coswara}~\citep{bhattacharya2023coswara}. We describe evaluation datasets in Appendix Table A3.

\paragraph{Evaluation}
21 unique downstream binary classification tasks across five datasets are leveraged to evaluate the quality of health acoustic representations generated from the learned audio encoders, including 13 human acoustic event classifications, five sleep apnea-specific tasks, and three cough relevant tasks. The cough tasks include COVID detection, sex classification, and smoking status classification.

For phases 1 and 2 of our study where we identify the best parameters for each augmentation, as well as the best combination of augmentations, we develop a composite score that aggregates performance across the various downstream tasks. The PSG, CoughVid, and Coswara datasets are segmented into two-second clips. For Flusense, we preprocess the data by segmenting variable length clips using the labeled timestamps. For FSD50K and Flusense, we adopt a lightweight evaluation strategy where we randomly sample a single two second long clip from each longer clip. We take the average area under the receiver operating characteristic curve (AUROC) across these tasks and use this composite measure to rank augmentation strategies.

For phase 3, we segment the PSG data into 10 second clips, and for FSD50K and Flusense, we crop or zero pad each clip to 10 seconds. We adopt a sliding window approach for FSD50K, Flusense, and PSG, where embeddings are generated for two-second windows with a step size of one second. We apply mean pooling to the resulting embeddings to generate our final output embedding.

For all phases, we use linear probing to evaluate the quality of the generated representations. We use logistic regression with cross-validated ridge penalty, which is trained to predict binary labels from the frozen precomputed embeddings (Köhn, 2015). We report AUROC for all tasks and use the DeLong method to compute the 95\% confidence intervals (CIs)~\citep{delong1988comparing}.

\paragraph{Baseline Models}
For comparative evaluation, we consider several off-the-shelf audio encoders, each trained on semantic or non-semantic speech data. Specifically, our baseline models include TRILL~\citep{shor2020towards}, which is a publicly available ResNet50 architecture trained on an AudioSet subset that is enriched with speech labels. FRILL~\citep{peplinski2020frill} is a light-weight MobileNet-based encoder distilled from TRILL. BigSSL-CAP12~\citep{shor2022universal} leverages a Conformer-based architecture, trained on YouTube and LibriLight.

\section{Results}

\paragraph{Optimal augmentation parameters}
In Appendix Table A1, we display the optimal parameters for each augmentation derived from the associated grid searches. We find that up to a certain threshold, generally more intense augmentation parameters yield better performance. 

\begin{figure}[h]
\centering
\includegraphics[width=1.0\linewidth]{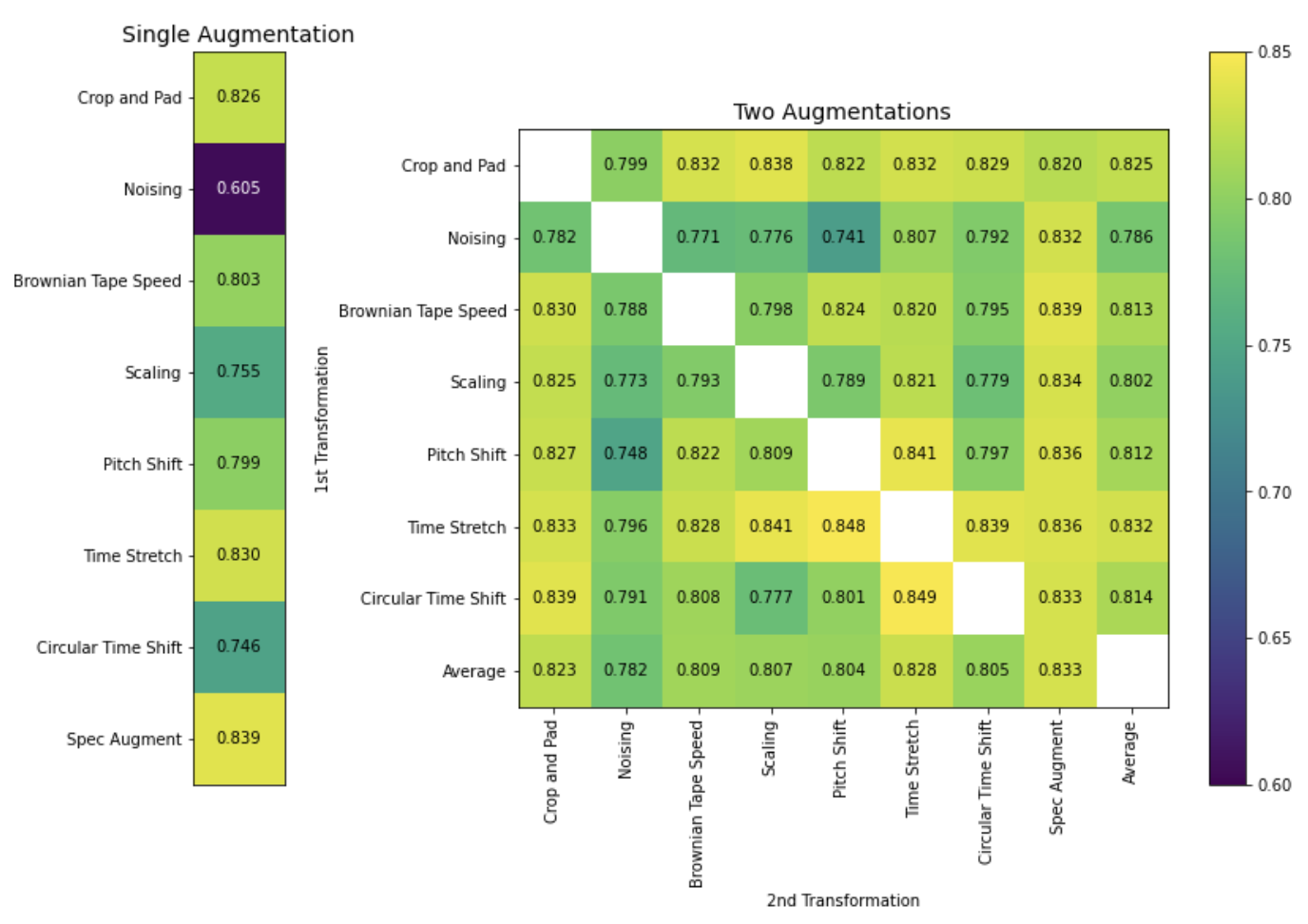} 
\caption{Evaluation performance for comparing augmentation combinations. (Left) from single augmentations. (Right) two augmentations applied where rows represent the first augmentation and columns represent the second augmentation.}
\label{fig2}
\end{figure}

\paragraph{Comparing augmentations}
Comparing the left and right panels of Figure~\ref{fig2} shows that many augmentations perform better in combination than individually. Our analysis indicates that the most effective single augmentation strategy is SpecAugment (left panel in Figure~\ref{fig2}). The most effective 2-step augmentation strategy involves applying circular time shift , followed by time stretch, as depicted in Figure~\ref{fig2}. Interestingly, circular time shift does not perform well on its own and each of these augmentations individually underperform SpecAugment. However, circular time shift and time stretch are synergistic when applied together. The right panel of Figure~\ref{fig2} shows that on average, time stretch is the most useful first augmentation, excluding SpecAugment which is always applied second or alone. SpecAugment is the most useful second augmentation on average.

\paragraph{Comparing to baselines}
Appendix Tables 4, 5 demonstrate performance of the best SimCLR model versus the baseline models on the validation set used for the comparison of augmentations. Overall, the performance of the SimCLR model is similar to BigSSL-CAP12, despite training on about 10x less hours of data and using a model that is nearly 10x smaller, and outperforms off-the-shelf audio encoders.

\section{Discussion and Conclusion}
We investigated a comprehensive list of augmentations for use in the health acoustic domain. We demonstrated the synergistic benefit of the circular time shift and time stretch augmentations. Circular time shift and time-stretching may synergistically improve model generalizability by introducing a diverse range of temporal patterns for the same sound.

There are few limitations worth noting. We decided to keep our test sets held out for ongoing model development, thus our comparisons to baselines may be optimistic. We also confined our analysis to a single Slowfast NFNet architecture. This leaves open the possibility that different architectures could yield varying results. Future research may focus on other augmentations, including frequency domain augmentations, as well as augmentations that better leverage health acoustic inductive biases. Additionally, incorporating labels during training~\citep{khosla2020supervised}, such as health signal type, may further improve the learned representations.

\acks{We thank Yun Liu from Google Research for his critical feedback, Shao-Po Ma for his preliminary work on the PSG dataset, CoughVID and Project Coswara teams for making the datasets publicly available, and the Google Research team for software and hardware infrastructure support. PSG, CoughVID and Coswara are licensed under a \href{https://creativecommons.org/licenses/by/4.0/}{Creative Commons Attribution 4.0 International (CC BY 4.0) License} and follow the Disclaimer of Warranties and Limitation of Liability in the license.}

% \clearpage
\bibliography{aug}

\clearpage

\newpage

\appendix
\setcounter{table}{0}
\renewcommand{\thetable}{A\arabic{table}}

\onecolumn
\section{SimCLR hyperparameters}
For training, we use 32 TPU-v3 cores with a batchsize of 4096. We use an AdamW optimizer with default parameters and a learning rate of 1.6e-3. We train all models for at least 300k steps, saving checkpoints every 5k steps. We select checkpoints that exhibit the best performance on the validation data after applying an exponential moving average, with a bias correction and a weight of 0.5, to the validation curves.

\section{Appendix Tables}

\begin{table*}[htbp]
\centering
\resizebox{\textwidth}{!}{
\begin{tabular}{cclll}
\toprule
\textbf{Augmentation} & \textbf{Apply On} & \multicolumn{1}{c}{\textbf{Description}} & \multicolumn{1}{c}{\textbf{Best Parameters}} & \multicolumn{1}{c}{\textbf{\begin{tabular}[c]{@{}c@{}}Grid Search \\ (Cartesian product of the lists)\end{tabular}}} \\
\midrule
Crop and pad & Temporal & \begin{tabular}[c]{@{}l@{}}Crops the audio signal and then\\ zero-pads to the input length.\end{tabular} & \begin{tabular}[c]{@{}l@{}}Probability = 1.0\\ Min fraction = 0.1\\ Max fraction = 0.5\end{tabular} & \begin{tabular}[c]{@{}l@{}}Probability = {[}0.8, 1.0{]}\\ Min fraction = {[}0.1, 0.3, 0.5{]}\\ Max fraction = {[}0.3, 0.5, 0.7{]}\\ Only when max fraction \textgreater min fraction\end{tabular}\\
\midrule
Noising & Temporal & Adds gaussian noise to the audio signal. & \begin{tabular}[c]{@{}l@{}}Probability = 1.0\\ Mean = 0.2\\ Stddev = 0.2\end{tabular} & \begin{tabular}[c]{@{}l@{}}Probability = {[}0.8, 1.0{]}\\ Mean = {[}-0.2, 0.0, 0.2{]}\\ Stddev = {[}0.2, 0.4, 0.6{]}\end{tabular} \\
\midrule
Brownian tape speed & Temporal & \begin{tabular}[c]{@{}l@{}}Simulates playing back the signal on a\\ tape while the playback speed at each time\\ step is drawn from a normal distribution.\end{tabular} & \begin{tabular}[c]{@{}l@{}}Probability = 0.8\\ Magnitude = 20\end{tabular} & \begin{tabular}[c]{@{}l@{}}Probability = {[}0.8, 1.0{]}\\ Magnitude = {[}2, 10, 20{]}\end{tabular} \\
\midrule
Scaling & Temporal & Modifies the audio gain. & \begin{tabular}[c]{@{}l@{}}Probability = 0.8\\ Min factor = 0.25\\ Max factor = 1.75\end{tabular} & \begin{tabular}[c]{@{}l@{}}Probability = {[}0.8, 1.0{]}\\ Min factor = {[}0.25, 0.75, 1.25{]}\\ Max factor = {[}0.75, 1.25, 1.75{]}\\ Only when max factor \textgreater min factor\end{tabular} \\
\midrule
Pitch shift & Temporal & \begin{tabular}[c]{@{}l@{}}Moves the pitch of the audio up or down\\ without changing its speed.\end{tabular} & \begin{tabular}[c]{@{}l@{}}Probability = 0.8\\ Min factor = 1.25\\ Max factor = 1.75\end{tabular} & \begin{tabular}[c]{@{}l@{}}Probability = {[}0.8, 1.0{]}\\ Min factor = {[}0.25, 0.75, 1.25{]}\\ Max factor = {[}0.75, 1.25, 1.75{]}\\ Only when max factor \textgreater min factor\end{tabular} \\
\midrule
Time stretch & Temporal & \begin{tabular}[c]{@{}l@{}}Slows and speeds up the audio signal\\ without changing its pitch.\end{tabular} & \begin{tabular}[c]{@{}l@{}}Probability = 0.8\\ Min time stretch = 0.75\\ Max time stretch = 1.75\end{tabular} & \begin{tabular}[c]{@{}l@{}}Probability = {[}0.8, 1.0{]}\\ Min time = {[}0.25, 0.75, 1.25{]}\\ Max stretch = {[}0.75, 1.25, 1.75{]}\\ Only when max factor \textgreater min factor\end{tabular} \\
\midrule
Circular time shift & Temporal & \begin{tabular}[c]{@{}l@{}}Translates the audio signal temporally\\ without truncating the signal,\\ while wrapping along the time axis.\end{tabular} & Probability = 1.0 & Probability = {[}0.8, 1.0{]} \\
\midrule
SpecAugment & Spectrogram & \begin{tabular}[c]{@{}l@{}}Applies masking to the temporal and\\ frequency axes.\end{tabular} & \begin{tabular}[c]{@{}l@{}}Probability = 1.0\\ Time mask max frames = 24\\ Time mask count = 20\\ Frequency mask max bins = 20\\ Frequency mask count = 5\end{tabular} & \begin{tabular}[c]{@{}l@{}}Probability = {[}0.8, 1.0{]}\\ Time mask max frames = {[}24, 36{]}\\ Time mask count = {[}10, 20{]}\\ Frequency mask max bins = {[}10, 20{]}\\ Frequency mask count = {[}3, 5{]}\end{tabular} \\
\bottomrule
\end{tabular}
}
\caption{Description of the augmentation strategies used in the study.}
\end{table*}

\begin{table*}[htbp]
\centering
\begin{tabular}{cc}
\toprule
\textbf{Class}           & \textbf{Estimated \# Audio Clips} \\
\midrule
Cough           & 77,000,000               \\
Speech          & 65,100,000               \\
Laughing        & 77,240,000               \\
Throat clearing & 3,300,000                \\
Baby cough      & 800,000                  \\
Breathing       & 31,500,000  \\
\bottomrule
\end{tabular}
\caption{Number of YouTube audio clips used for training.}
\end{table*}

\begin{table*}[htbp]
\centering
\resizebox{\textwidth}{!}{
\begin{tabular}{ccccc}
\toprule
\textbf{Dataset} & \textbf{Tasks} & \textbf{\begin{tabular}[c]{@{}c@{}}Number of examples for\\ training linear probes\end{tabular}} & \textbf{\begin{tabular}[c]{@{}c@{}}Number of examples\\ for evaluation\end{tabular}} & \textbf{Reference} \\
\midrule
FSD50K & Health acoustic event (6 tasks) & 32,652 & 8,313 &~\cite{fonseca2021fsd50k} \\
\midrule
Flusense & Health acoustic event (7 tasks) & 7,537 & 1,779 &~\cite{al2020flusense} \\
\midrule
PSG & Apnea, arousal events (5 tasks) & 7,320 & 3,625 &~\cite{korompili2021psg} \\
\midrule
CoughVID & COVID, sex (2 tasks) & 44,249 & 15,083 &~\cite{orlandic2021coughvid} \\
\midrule
Coswara & \begin{tabular}[c]{@{}l@{}}COVID, sex, smoking status (3 tasks)\end{tabular} & 10,230 & 4,285 &~\cite{bhattacharya2023coswara} \\
\bottomrule
\end{tabular}
}
\caption{Evaluation datasets statistics.}
\end{table*}

\begin{table*}[htbp]
\centering
\resizebox{\textwidth}{!}{
\begin{tabular}{cccccc}
\toprule
\textbf{Dataset} & \textbf{Task} & \textbf{TRILL} & \textbf{FRILL} & \textbf{BigSSL-CAP12} & \textbf{SimCLR (ours)} \\
\midrule
FSD50K & Breathing & 0.973 (0.958, 0.988) & 0.974 (0.961, 0.987) & \textbf{0.983 (0.973, 0.993)} & 0.982 (0.969, 0.995) \\
 & Cough & 0.988 (0.982, 0.994) & 0.986 (0.979, 0.993) & 0.998 (0.996, 1) & \textbf{0.999 (0.998, 1)} \\
 & Laughter & 0.985 (0.978, 0.992) & 0.984 (0.977, 0.992) & \textbf{0.994 (0.988, 0.999)} & 0.991 (0.983, 1) \\
 & Sneeze & 0.913 (0.757, 1) & 0.960 (0.896, 1) & \textbf{0.997 (0.995, 1)} & 0.988 (0.969, 1) \\
 & Speech & 0.970 (0.958, 0.982) & 0.972 (0.962, 0.983) & \textbf{0.982 (0.974, 0.991)} & 0.978 (0.967, 0.988) \\
 & All Respiratory sounds & 0.978 (0.972, 0.985) & 0.979 (0.973, 0.985) & 0.985 (0.977, 0.994) & \textbf{0.990 (0.984, 0.995)} \\
\midrule
Flusense & Breathe & 0.732 (0.602, 0.861) & 0.741 (0.614, 0.869) & 0.769 (0.636, 0.902) & \textbf{0.816 (0.706, 0.925)} \\
 & Cough & 0.656 (0.614, 0.698) & 0.658 (0.616, 0.700) & 0.675 (0.635, 0.716) & \textbf{0.703 (0.662, 0.743)} \\
 & Gasp & 0.731 (0.624, 0.837) & 0.721 (0.618, 0.824) & 0.766 (0.676, 0.855) & \textbf{0.777 (0.675, 0.880)} \\
 & Sneeze & 0.719 (0.663, 0.776) & 0.717 (0.66, 0.773) & 0.780 (0.731, 0.829) & \textbf{0.789 (0.740, 0.838)} \\
 & Sniffle & 0.734 (0.671, 0.798) & 0.727 (0.662, 0.791) & 0.717 (0.648, 0.787) & \textbf{0.762 (0.697, 0.827)} \\
 & Speech & 0.711 (0.670, 0.751) & 0.701 (0.659, 0.742) & \textbf{0.764 (0.724, 0.804)} & 0.751 (0.715, 0.788) \\
 & Throat clearing & 0.811 (0.692, 0.931) & 0.756 (0.620, 0.891) & \textbf{0.914 (0.863, 0.964)} & 0.788 (0.671, 0.905) \\
\midrule
PSG & OSA & 0.681 (0.643, 0.720) & 0.697 (0.658, 0.736) & \textbf{0.770 (0.735, 0.806)} & 0.700 (0.663, 0.738) \\
 & Central & 0.640 (0.441, 0.838) & 0.695 (0.537, 0.852) & \textbf{0.725 (0.553, 0.896)} & 0.690 (0.510, 0.870) \\
 & Mixed & 0.728 (0.658, 0.797) & 0.732 (0.663, 0.800) & \textbf{0.788 (0.726, 0.850)} & 0.719 (0.654, 0.783) \\
 & Hypopnea & 0.497 (0.445, 0.549) & 0.537 (0.485, 0.588) & \textbf{0.639 (0.590, 0.688)} & 0.549 (0.500, 0.597) \\
 & Arousal & 0.716 (0.674, 0.759) & 0.732 (0.691, 0.772) & 0.728 (0.686, 0.770) & \textbf{0.784 (0.746, 0.822)} \\
\bottomrule
\end{tabular}
}
\caption{Performance comparison (AUROC with 95\% confidence intervals) on downstream tasks in FSD50K, Flusense and PSG datasets. OSA: obstructive sleep apnea.}
% TODO(ckbjimmy): ensure 3dp
\end{table*}

\begin{table*}[htbp]
\centering
\resizebox{\textwidth}{!}{
\begin{tabular}{cccccc}
\toprule
\textbf{Task} & \textbf{Dataset} & \textbf{TRILL} & \textbf{FRILL} & \textbf{BigSSL-CAP12} & \textbf{SimCLR (ours)} \\
\midrule
COVID & CoughVID & 0.613 (0.592, 0.634) & 0.611 (0.59, 0.632) & 0.621 (0.6, 0.642) & \textbf{0.622 (0.601, 0.643)} \\
 & Coswara & 0.573 (0.54, 0.607) & 0.591 (0.557, 0.625) & 0.597 (0.565, 0.628) & \textbf{0.769 (0.752, 0.785)} \\
\midrule
Smoker & Coswara & 0.62 (0.589, 0.651) & 0.579 (0.548, 0.609) & \textbf{0.624 (0.594, 0.654)} & 0.591 (0.560, 0.621) \\
\midrule
Sex & CoughVID & 0.839 (0.831, 0.847) & 0.83 (0.822, 0.838) & 0.847 (0.839, 0.855) & \textbf{0.862 (0.854, 0.869)} \\
 & Coswara & 0.872 (0.861, 0.883) & 0.827 (0.814, 0.84) & 0.900 (0.890, 0.910) & \textbf{0.903 (0.893, 0.913)} \\
\bottomrule
\end{tabular}
}
\caption{Performance comparison (AUROC with 95\% confidence intervals) on downstream tasks in CoughVID and Coswara datasets.}
% TODO(ckbjimmy): ensure 3dp
\end{table*}

\end{document}